# Do They All Look the Same? Deciphering Chinese, Japanese and Koreans by Fine-Grained Deep Learning

Yu Wang, Haofu Liao, Yang Feng, Xiangyang Xu, Jiebo Luo*
Department of Computer Science
University of Rochester
Rochester, NY, USA

## Abstract

We study to what extend Chinese, Japanese and Korean faces can be classified and which facial attributes offer the most important cues. First, we propose a novel way of obtaining large numbers of facial images with nationality labels. Then we train state-of-the-art neural networks with these labeled images. We are able to achieve an accuracy of 75.03% in the classification task, with chances being 33.33% and human accuracy 38.89% . Further, we train multiple facial attribute classifiers to identify the most distinctive features for each group. We find that Chinese, Japanese and Koreans do exhibit substantial differences in certain attributes, such as bangs, smiling, and bushy eyebrows. Along the way, we uncover several gender-related cross-country patterns as well. Our work, which complements existing APIs such as Microsoft Cognitive Services and Face++, could find potential applications in tourism, e-commerce, social media marketing, criminal justice and even counter-terrorism.

## Introduction

China, Japan and Korea are three of the world's largest economies, ranking 2nd, 3rd and 11th respectively.[1] Each country also boasts a large population: China ranks 1st in the world, Japan 11th, and South Korea 27th.[2] Given their geographical proximity, the three countries also share strong economic ties with each other, through trade and tourism, for example.

Thanks to their large economies, large populations, their geographical proximity, and the fact that they do look very similar, there is the widely shared impression that these three peoples really look the same. Some suggest that the main differences derive from mannerism and fashion.[3] There also exist quite a few websites that put up Asian face classification challenges, which only helps reinforce the wisdom that Chinese, Japanese and Koreans all look the same.[4]

Recent advances in computer vision (Krizhevsky, Sutskever, and Hinton 2012a; Simonyan and Zisserman 2015; He et al. 2015), on the other hand, have made object detection and classification increasingly accurate. In particular, face detection and race classification (Farfade, Saberian, and Li 2015; Fu, He, and Hou 2014; Wang, Li, and Luo 2016) have both achieved very high accuracy, largely thanks to the adoption of deep learning (LeCun, Bengio, and Hinton 2015) and the availability of large datasets (Huang et al. 2007; Jr. and Tesafaye 2006; Phillips et al. 1998) and more recently (Guo et al. 2016).

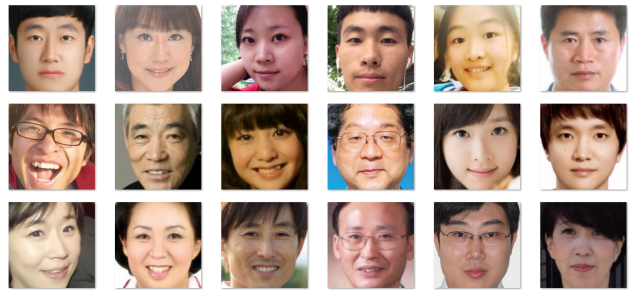

Figure 1: When randomly shuffled, Chinese, Japanese, and Koreans are difficult to distinguish.

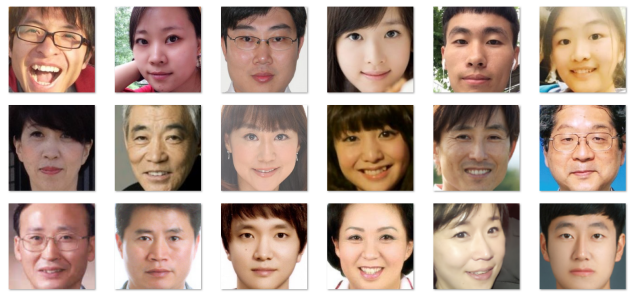

Figure 2: When grouped together, several patterns seem to emerge, thus the power of big data. Top role: Chinese, Middle: Japanese, Bottom: Korean.

In this paper, we attempt to break the conventional wisdom that "they all look the same" using big data computer

---

Jiebo Luo (jluo@cs.rochester.edu) is the corresponding author of this paper. All rights reserved.

[1] http://databank.worldbank.org/data/download/GDP.pdf.
[2] http://databank.worldbank.org/data/download/POP.pdf.
[3] See for example, "Japanese, Korean, Chinese. What's the difference?" https://blog.gaijinpot.com/japanese-korean-chinese.
[4] Two prominent examples are alllooksame.com and asianlookalikes.com.

vision. We collect 39,883 labeled faces (roughly 1: 1: 1) from Twitter (Figure 1). We use our images to fine-tune VGG (Simonyan and Zisserman 2015) and ResNet (He et al. 2015), the two state-of-the-art architectures. With Resnet, we achieve a remarkable accuracy of 75.03%, with chances being 33.33% and human performance at 38.89%.[5]

Moreover, we further classify the facial attributes of these three peoples in an effort to identify fine-grained patterns. Our study shows, for example, Chinese are most likely to have bushy eyebrows, Japanese smile the most, and Koreans are most likely to have black hair. We also briefly report on cross-country gender differentials.

## Related Literature

Recent advances in computer vision (Krizhevsky, Sutskever, and Hinton 2012a; Simonyan and Zisserman 2015; He et al. 2015) have made object detection and classification increasingly accurate. In particular, face detection, gender and race classification (Farfade, Saberian, and Li 2015; Levi and Hassner 2015; Fu, He, and Hou 2014; Wang, Li, and Luo 2016) have both achieved very high accuracy, largely thanks to the adoption of deep learning (LeCun, Bengio, and Hinton 2015) and the availability of large datasets (Huang et al. 2007; Jr. and Tesafaye 2006; Phillips et al. 1998) and more recently (Guo et al. 2016). It has been empirically observed that deeper neural networks tend to outperform shall networks. As a result, quite a few very deep learning architectures have been proposed (Krizhevsky, Sutskever, and Hinton 2012a; Romero et al. 2015; Simonyan and Zisserman 2015; He et al. 2015) and more recently recursive highway networks (Zilly et al. 2016), which is considered as a generalization of Resnet.

As a subfield of image classification, facial attribute classification has also been an area of active research. One of the early works in this area is (Belhumeur and Nayar 2008), which uses SVM for classification and adaboost for feature selection to study 10 facial attributes. Following (Belhumeur and Nayar 2008), (Kumar et al. 2009) attempts to classify 65 attributes with which to perform face verification. Methodologically, (Kumar et al. 2009) differs from (Belhumeur and Nayar 2008) in that it uses forward selection instead of adaboost. (Zhang et al. 2014) is one of the early works that apply deep learning to facial attribute classification. A noteworthy feature is that all the attributes share the same parameters until the last layer which is attribute-specific and is subsequently used as input for logistic regression. (Liu et al. 2015), which has achieved state-of-the-art performance, trains two networks, the first one for face localization and the second for attribute classification. Our work follows (Liu et al. 2015), and uses the same dataset as theirs, except that (1) for simplicity we use OpenCV to locate faces and (2) for accuracy we train a separate neural network for each attribute.

In seeking to identify the most distinctive cues, our work is related to (Kumar et al. 2009) and (Doersch et al. 2012). The former uses facial attributes for identity verification and the latter tries to identify architectural features that distinguish Paris from other cities such as Boston and London.

[5]alllooksame.com.

## Data Collection and Pre-processing

We have two main data sources: Twitter and the CelebA dataset.[6] We derive from Twitter the labeled Chinese, Japanese and Korean images, which are later used as input to the Resnet. We use CelebA to train the facial attribute classifiers. These classifiers are then used to classify the labeled Twitter images.

### Twitter Images

We collect profile images of the Twitter followers of Asian celebrities. For the Chinese celebrity, we choose Kai-Fu Lee (1.64 million followers). For the Japanese celebrity, we choose the Japanese Prime Minister Shinzo Abe (605,000 followers). For the Korean celebrity, we choose the South Korean president Park Geun Hye (409,000 followers).

Not all followers carry the right label. For example, a Japanese might be following the Korean president. To solve this problem, we restrict the selection to followers that speak the label language (e.g. Chinese for Chinese followers). We detect followers' language from their user name and self-description. Then we collect profile images that meet our language restriction.

To process the profile images, we first use OpenCV to identify faces (Jia and Cristianini 2015), as the majority of profile images only contain a face.[7] We discard images that do not contain a face and the ones in which OpenCV is not able to detect a face. When multiple faces are available, we keep all of them, assuming only individuals of the same nationality are present. Out of all facial images thus obtained, we select only the large ones. Here we set the threshold to 12kb. This ensures high image quality and also helps remove empty faces. Lastly we resize those images to (224, 224, 3) (Krizhevsky, Sutskever, and Hinton 2012a). Eventually, we get 39,883 images, after cleaning. In Table 1, we report the summary statistics of the images used in the experiments.

Table 1: Summary Statistics of the Asian Face Dataset

|          | Chinese | Japanese | Korean |
|----------|---------|----------|--------|
| # Images | 13,429  | 12,914   | 13,540 |

### CelebA Images

The CelebA dataset contains 202,599 images taken from ten thousand individuals. To ensure no overlapping betwwen training and testing, (Liu et al. 2015) uses images from the first eight thousand individuals for training the neural networks, images for the ninth thousand individuals for training SVM, and images from the last thousand identities for testing.

In our work, we follow this practice in dividing the training, developing, and testing dataset. We use OpenCV to locate faces in each subset and eventually have 148,829 images for training, 18,255 images for development, and 18,374 images for testing.

[6]http://mmlab.ie.cuhk.edu.hk/projects/CelebA.html.
[7]http://opencv.org.

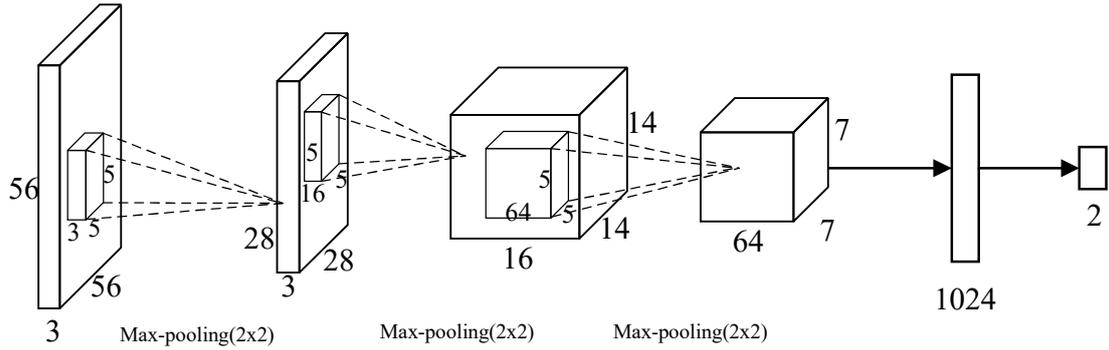

Figure 3: Architecture of the neural net for attribute classification, designed the same for all 40 classifiers.

## Experiments

We conduct two experiments. In the first experiment, we use the labeled Twitter images to finetune the Resnet and investigate to what extent Chinese, Japanese and Koreans can be classified. In our second experiment, we train 40 facial attribute classifiers and examine which attributes contain the most important cues in distinguishing the three groups.

### Face Classification

We split our dataset into development set, validation set, and test set ( 8: 1: 1) and experiment with different architectures from shallow networks (3-5 layers) to the 16-layer VGG and 50-layer ResNet. In our experiments, all networks would converge (Figure 4), but we observe that as the network gets deeper, we are able to achieve better results (Srivastava, Greff, and Schmidhuber 2015), from 60% accuracy with shallow networks to an overall accuracy of 75.03% with Resnet.

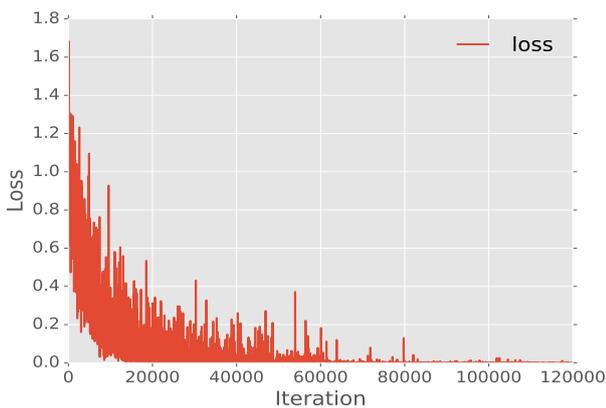

Figure 4: Convergence of the ResNet during the training.

In Table 2, we report the confusion matrix for the testing images. Note that all the three peoples look equally "confusing" to the computer: the off-diagonal elements are roughly equal. The result we achieve answers in a definitive manner that Chinese, Japanese and Koreans are distinguishable. But it also suggests that this is a challenging task, which leads to our experiment on facial attribute classification.

Table 2: ResNet: Confusion Matrix for Asian Face Classification

| Total=3,988 | | Prediction | | |
|---|---|---|---|---|
| | | Chinese | Japanese | Korean |
| Truth | Chinese (1343) | 78.21% | 11.86% | 9.93% |
| | Japanese (1291) | 13.33% | 72.80% | 13.87% |
| | Korean (1354) | 13.88% | 12.33% | 73.80% |

*Numbers do not necessarily add up to 1 due to rounding.

### Attribute Classification

In this experiment, our goal is to examine which facial attributes offer the most important cue in distinguishing the three peoples. In order to do so, we first build attribute classifiers, as detailed in the 3rd section, and use these classifiers to classify Twitter images for each country group.

We construct a separate neural network for each of the 40 attributes in the CelebA dataset. The neural nets all share the same structure (Figure 3) but do not share any parameters as is the case in (Liu et al. 2015). Consequently, we calculate the cross-entropy loss function separately, instead of as a sum, as L=$y_i$log$p(y_i$ |x)+(1-$y_i$)log(1-$p(y_i$ |x)) for each attribute $i$, where $p(y_i = 1|x) = \frac{1}{1+exp(-f(x))}$ and x represents the facial image.

The performance of the neural nets are reported in Table 3. It should be noted that while the results from (Liu et al. 2015) are also reported, the results are not strictly comparable as we are not using exactly the same images. The comparison serves to demonstrate that our classifiers are sufficient for performing attribute classification on our Twitter images, which is the focus of our work.

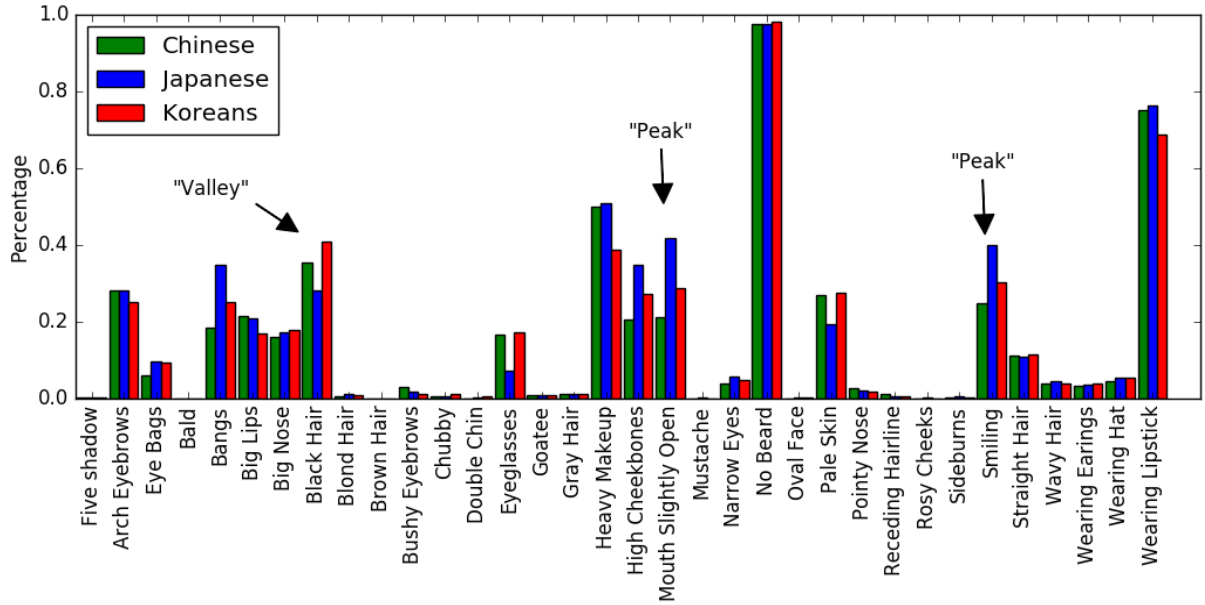

Figure 5: Attribute comparison of Chinese, Japanese and Korean females.

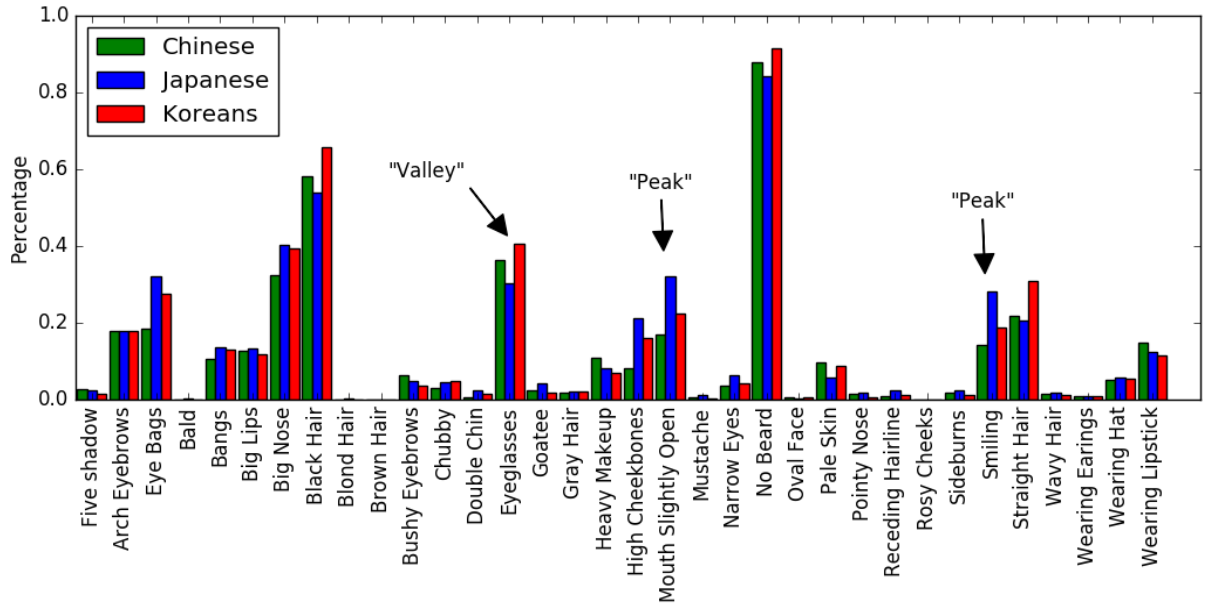

Figure 6: Attribute comparison of Chinese, Japanese and Korean males.

In Figures 5 and 6, we report the percentage of individuals that possess the corresponding facial attributes. Several patterns, which hold across gender, immediately suggest themselves:

Table 3: Attribute Classification Results

| Attribute | LNets+ANet | Our Result |
|---|---|---|
| Five Shadow | 91% | **91.41%** |
| Arch Eyebrows | 79% | 76.78% |
| Attractive | 81% | 72.82% |
| Bags Under Eyes | 79% | **79.54%** |
| Bald | 98% | 97.93% |
| Bangs | 95% | 91.80% |
| Big Lips | 68% | 67.29% |
| Big Nose | 78% | 76.57% |
| Black Hair | 88% | 80.25% |
| Blond Hair | 95% | 92.45% |
| Blurry | 84% | **95.43%** |
| Brown Hair | 80% | **81.31%** |
| Bushy Eyebrows | 90% | 89.26% |
| Chubby | 91% | **94.47%** |
| Double Chin | 92% | **95.33%** |
| Eyeglasses | 99% | 98.11% |
| Goatee | 95% | **95.89%** |
| Gray Hair | 97% | 96.54% |
| Heavy Makeup | 90% | 85.17% |
| High Cheekbones | 87% | 84.14% |
| Male | 98% | 95.16% |
| Mouth Slightly Open | 92% | 91.02% |
| Mustache | 95% | **96.66%** |
| Narrow Eyes | 81% | **85.91%** |
| No Beard | 95% | 93.07% |
| Oval Face | 66% | **69.28%** |
| Pale Skin | 91% | **93.08%** |
| Pointy Nose | 72% | **72.25%** |
| Receding Hairline | 89% | **91.70%** |
| Rosy Cheeks | 90% | **92.67%** |
| Sideburns | 96% | **96.11%** |
| Smiling | 92% | 89.53% |
| Straight Hair | 73% | **77.19%** |
| Wavy hair | 80% | 70.32% |
| Wearing Earrings | 82% | **84.87%** |
| Wearing Hat | 99% | 98.24% |
| Wearing Lipstick | 93% | 89.75% |
| Wearing Necklace | 71% | **85.74%** |
| Wearing Necktie | 93% | **93.13%** |
| Young | 87% | 83.70% |
| *Average* | 87% | **87.30%** |

Note: Equivalent or better performances are marked bold.

1. Bangs are most popular among Japanese and least popular among Chinese.
2. Japanese smile the most and Chinese the least.
3. Japanese have the most eyebags, followed by Koreans.
4. Chinese are the most likely to have bushy eyebrows.
5. Koreans are the mostly likely to have black hair and Japanese are the least likely.

## Cross-Country Gender Differentials

While our paper focuses on country comparisons, we also briefly summarize some of the significant findings on cross-country gender differentials that are either cultural or social in nature.

First, we observe that in all three countries under study females tend to smile more than males (Figure 7). In (Ginosar et al. 2015), after analyzing decades of high school yearbooks in the U.S., the authors conclude that smiles are increasing over time and women have always been smiling more than men. Our finding can be considered as the Asian counterpart to that observation, and we suggest that this might be caused by social norms.

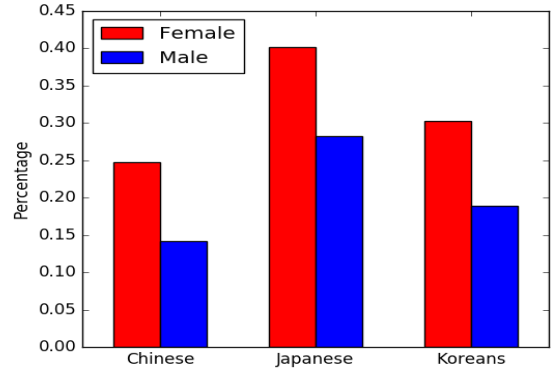

Figure 7: Cross-country gender differentials: smiling.

Second, we notice that wearing glasses is very common in all three countries and that men are twice more likely to be wearing glasses than women in their Twitter profiles. While we do not have definitive interpretations, we suggest that this might be caused by work-related pressure, which is consistent with our finding that men are also more likely to have eye bags.[8]

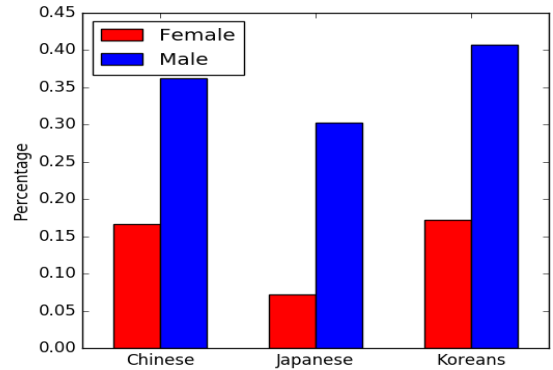

Figure 8: Cross-country gender differentials: eyeglasses.

## Limitations

Our work is built on the assumption that Twitter users, celebrity followers in particular, are representative of the demographics of the entire population. This assumption may not exactly hold as various demographic dimensions such as gender and age are skewed in Twitter (Mislove et al. 2011). In particular, followers of Kai-fu Lee, Shinzo Abe and Park Geun Hye could differ from the general population and this could produce selection bias and affect the magnitude of our estimation (Heckman 1979). Nonetheless, we believe the direction of our estimates will remain consistent, as several of our findings are confirmed by social stereotypes and research on other regions. Also, this concern could be alleviated to some extent by examining several other celebrities.

---
[8]Related graphs can be found at the authors' official website.

## Conclusion

In this paper, we have demonstrated that Chinese, Japanese and Koreans do look different. By assembling a large data set of labeled images and experimenting with different neural network architectures, we have achieved a remarkable accuracy 75.03%, almost twice as high as the human average accuracy. We believe an even higher accuracy is achievable with a larger and cleaner dataset. Towards that goal, we are experimenting with neural network face detectors (Farfade, Saberian, and Li 2015) specifically designed for Twitter images.

We have also examined 40 facial attributes of the three populations in an effort to identify the important cues that assist classification. Our study has shown that Chinese, Japanese and Koreans do differ in several dimensions but overall are very similar. Along the way, we have also uncovered quite a few interesting gender related phenomena prevalent in East Asia.

Our work, which complements existing APIs such as Microsoft Cognitive Services and Face++, could find wide applications in tourism, e-commerce, social media marketing, criminal justice and even counter-terrorism.